# Total Variation and Euler's Elastica for Supervised Learning

Tong Lin                                              tonglin123@gmail.com
Hanlin Xue                                            xuehl@cis.pku.edu.cn
Ling Wang*                                            ling.wang.nj@gmail.com
Hongbin Zha                                           zha@cis.pku.edu.cn

The Key Laboratory of Machine Perception (Ministry of Education), Peking University, Beijing, China
*LTCI, Télécom ParisTech, Paris, France

## Abstract

In recent years, total variation (TV) and Euler's elastica (EE) have been successfully applied to image processing tasks such as denoising and inpainting. This paper investigates how to extend TV and EE to the supervised learning settings on high dimensional data. The supervised learning problem can be formulated as an energy functional minimization under Tikhonov regularization scheme, where the energy is composed of a squared loss and a total variation smoothing (or Euler's elastica smoothing). Its solution via variational principles leads to an Euler-Lagrange PDE. However, the PDE is always high-dimensional and cannot be directly solved by common methods. Instead, radial basis functions are utilized to approximate the target function, reducing the problem to finding the linear coefficients of basis functions. We apply the proposed methods to supervised learning tasks (including binary classification, multi-class classification, and regression) on benchmark data sets. Extensive experiments have demonstrated promising results of the proposed methods.

## 1. Introduction

Supervised learning (Bishop, 2006; Hastie T., 2009) infers a function that maps inputs to desired outputs under the guidance of training data. Two main tasks in supervised learning are classification and regression. A huge number of supervised learning methods have been developed in several decades (see a com-prehensive empirical comparison of these methods in (Caruana & Niculescu-Mizil, 2006)). Existing methods can be roughly divided into statistics based and function learning based (Kotsiantis et al., 2006). One advantage of function learning methods is that powerful mathematical theories in functional analysis can be utilized rather than doing optimizations on discrete data points.

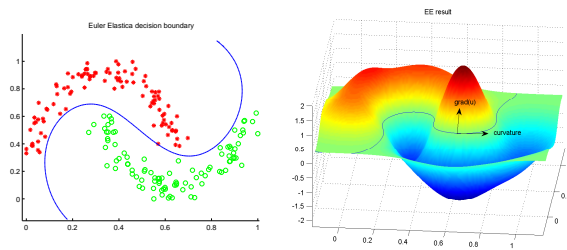

*Figure 1.* Results on two moon data by the EE classifier: (Left) decision boundary; (Right) learned target function.

Most function learning methods can be derived from Tikhonov regularization, which minimizes a loss term plus a smoothing regularizer. The most successful classification and regression method is SVM (Bishop, 2006; Hastie T., 2009; Shawe-Taylor & Cristianini, 2000), whose cost function is composed of a hinge loss and a RKHS norm determined by a kernel. Replacing the hinge loss by a squared loss, the modified algorithm is called Regularized Least Squares (RLS) method (Rifkin, 2002). In addition, manifold regularization (Belkin et al., 2006) introduced a regularizer of squared gradient magnitude on manifolds. Its discrete version amounts to graph Laplacian regularization (Nadler et al., 2009; Zhou & Schölkopf, 2005), which approximates the original energy functional. A most recent work is the geometric level set (GLS) classifier (Varshney & Willsky, 2010), with an energy functional composed of a margin-based loss and a geometric regularization term based on the surface area of the decision boundary. Experiments showed that GLS is competitive with SVM and other state-of-the-art classifiers.

In this paper, the supervised learning problem is formulated as an energy functional minimization under Tikhonov regularization scheme, with the energy composed of a squared loss and a total variation (TV) penalty or an Euler's elastica (EE) penalty. Since the TV and EE models have achieved great success in image denoising and image inpainting (Aubert & Kornprobst, 2006; Barbero & Sra, 2011; Chan & Shen, 2005), a natural question is whether the success of TV and EE models on image processing applications can be transferred to high dimensional data analysis such as supervised learning. This paper investigates the question by extending TV and EE models to supervised learning settings, and evaluating their performance on benchmark data sets against state-of-the-art methods. Figure 1 shows the classification result on the popular two moon data by the EE classifier, and the learned target function. Interestingly, the GLS classifier (Varshney & Willsky, 2010) is also motivated by image processing techniques, and its gradient descent time marching leads to a mean curvature flow.

The paper is organized as follows. We begin with a brief review of TV and EE in Section 2. In Section 3 the proposed models are described, and numerical solutions are developed in Section 4. Section 5 presents the experimental results, and Section 6 concludes this paper.

## 2. Preliminaries

We briefly introduce total variation and Euler's elastica from an image processing perspective, and point out connections with prior work in the machine learning literature.

### 2.1. Total Variation (TV)

The total variation of a 1D real-valued function $f$ is defined as

$$V_b^a(f) = \sup \sum_{i=0}^{n_p-1} |f(x_{i+1}) - f(x_i)|,$$

where the supremum runs over all partitions of given interval $[a, b]$. If $f$ is differentiable, the total variation can be written as

$$V_b^a(f) = \int_a^b |f'(x)| dx.$$

Simply, it is a measure of the total quantity of the change of a function. Notice that if $f'(x) > 0, x \in [a, b]$, it is exactly $f(b) - f(a)$ by the basic theorem of calculus. Total variation has been widely used for image processing tasks such as denoising and inpainting. The pioneering work is Rudin, Osher, and Fatemi's image denoising model (Rudin et al., 1992):

$$J = \int_\Omega ((I - I_0)^2 + \lambda |\nabla I|) dx,$$

where $I_0$ is the input image with noise, $I$ the desired output image, $\lambda$ a regulation parameter that balances two terms, and $\Omega$ a $2D$ image domain. The first fitting term measures the fidelity to the input, while the second is a $p$-Sobolev regularization term ($p = 1$) where $\nabla I$ is understood in the distributional sense. The main merit is to preserve significant image edges during denoising (Aubert & Kornprobst, 2006; Chan & Shen, 2005). Note that TV may have different definitions (Barbero & Sra, 2011).

In the machine learning literature, $p$-Sobolev regularizer can be found in nonparametric smoothing splines, generalized additive models, and projection pursuit regression models (Hastie T., 2009). Specifically, Belkin et al. proposed the manifold regularization term

$$\int_{x \in M} |\nabla_M f|^2 dx,$$

on a manifold M (Belkin et al., 2006). On the other hand, discrete graph Laplacian regularization was discussed in (Zhou & Schölkopf, 2005) as

$$\sum_{v \in V} |\nabla_v f|^p,$$

where $v$ is a vertex from $V$, and $p$ is an arbitrary number. This penalty measures the roughness of $f$ over a graph.

### 2.2. Euler's Elastica (EE)

Euler (1744) first introduced the elastica energy for a curve on modeling torsion-free elastic rods. Then Mumford (Mumford, 1991) reintroduced elastica into computer vision. Later, elastica based image inpainting methods were developed in (Chan et al., 2002; Masnou & Morel, 1998).

A curve $\gamma$ is said to be Euler's elastica if it is the equilibrium curve of the elasticity energy:

$$E[\gamma] = \int_\gamma (a + b\kappa^2) ds, \qquad (1)$$

where $a$ and $b$ stand for two positive constant weights, $\kappa$ denotes the scalar curvature, and $ds$ is the arc length element. Euler obtained the energy in studying the steady shape of a thin and torsion-free rod under external forces. The curve implies the lowest elastica energy, thus getting its name. According to (Mumford, 1991), the key link between the elastica and image inpainting relies on the the interpolation capability of elastica. That is, elastica can comply to the connectivity principle better than total variation. Such kinds of "nonlinear splines", like classical polynomial splines, are natural tools for completing the missing or occluded edges.

The Euler's elastica based inpainting model was proposed as (Chan & Shen, 2005)

$$J = \int_{\Omega \setminus D} (I - I_0)^2 dx + \lambda \int_{\Omega} (a + b\kappa^2) |\nabla I| dx, \quad (2)$$

where $D$ is the region to be inpainted, $\Omega$ the whole image domain, and $\kappa$ the curvature of the associated level set curve with

$$\kappa = \nabla \cdot \left( \frac{\nabla I}{|\nabla I|} \right). \quad (3)$$

By using calculus of variation, its minimization is reduced to an nonlinear Euler-Lagrange equation. The finite difference scheme can be used to give numerical implementation, and experimental results show that the EE based inpainting performs better than its TV version.

Elastica can be regarded as an extension of total variation, since elastica degenerates to total variation if set $a = 1$ and $b = 0$. In fact, elastica is a combination of total variation suppressing oscillations in the gradient direction, and a curvature regularization term that penalizes non-smooth level set curves (see Figure 1).

## 3. The Proposed Framework

### 3.1. Problem Setup

The general supervised learning problem can be described as follows: given a training data $\{(\mathbf{x}_1, y_1), ...(\mathbf{x}_n, y_n)\}$ with data points $\mathbf{x}_i \in \Omega \subset R^d$ and corresponding target varibles $y_i$, the goal is to estimate an unknown function $u(\mathbf{x})$ for a new point $\mathbf{x}$. The difference between classification and regression lies only in the corresponding target values, with one discrete and the other continuous. The widely used Tikhonov regularization framework for supervised learning can be formulated as:

$$\min \sum_{i=1}^{n} L(u(\mathbf{x}_i), y_i) + \lambda S(u), \quad (4)$$

where $L$ denotes a loss function and $S(u)$ is a smoothing term. A variety of loss functions $L$ have been proposed in the literature: hinge loss for SVM, squared loss for RLS, logistic loss for logistic regression, Huber loss, exponential loss, and among others. Throughout the paper, squared loss is used in all models due to its rather simpler differential form.

### 3.2. Laplacian Regularization (LR)

A commonly used model using squared loss can be written as

$$\min \sum_{i=1}^{n} (u(\mathbf{x}_i) - y_i)^2 + \lambda S(u). \quad (5)$$

If the RKHS norm is used for the smoothing term, the model is called regularized least squares (RLS) (Rifkin, 2002). Another natural choice is the squared $L_2$-norm of the gradient: $S(u) = |\nabla u|^2$, as proposed in (Belkin et al., 2006). Under a continuous setting, we get the following Laplacian regularization (LR) model:

$$J_{LR}[u] = \int_{\Omega} ((u - y)^2 + \lambda |\nabla u|^2) d\mathbf{x}. \quad (6)$$

This LR model has been widely used in the image processing literatures. Using calculus of variations, the minimization can be reduced to the following Euler-Lagrange partial differential equation (PDE) with a natural boundary condition along $\partial \Omega$:

$$\begin{cases} -\lambda \triangle u + (u - y) = 0 \\ \frac{\partial u}{\partial \mathbf{n}}|_{\partial \Omega} = 0 \end{cases}, \quad (7)$$

where $\triangle u$ is the Laplacian of $u$, and $\mathbf{n}$ denotes the normal vector along the boundary $\partial \Omega$. This PDE is relatively simple and can be solved using common methods in two and three dimensions. The next Section provides a function approximation method for solving the PDE in high dimensions.

### 3.3. Total Variation (TV) based Smoothing

Our goal is to explore how TV and EE can be applied to classification and regression problems on high dimensional data sets. A typical procedure has three steps: (a) Set the function learning problem under a continuous setting and design a proper energy functional; (b) Derive the Euler-Lagrange PDE via the calculus of variations; (c) Solve the PDE on discrete data points.

Similar to image denoising, total variation (TV) based supervised learning can be formulated as

$$J_{TV}[u] = \int_{\Omega} (u - y)^2 d\mathbf{x} + \lambda \int_{\Omega} |\nabla u| d\mathbf{x}. \quad (8)$$

Note that for binary classification the zero level set of $u$ serves as the final decision boundary. The only difference between LR and TV is just the $p$-Sobolev regularizer with $p = 2$ for LR and $p = 1$ for TV. Intuitively, LR penalizes too much gradients on edges, while TV can permit sharper edges near the decision boundaries between two classes.

### 3.4. Euler's Elastica (EE) based Smoothing

Elastica based supervised learning can be formulated as

$$J_{EE}[u] = \int_\Omega (u-y)^2 d\mathbf{x} + \lambda \int_\Omega (a+b\kappa^2)|\nabla u|d\mathbf{x}, \quad (9)$$

where

$$\kappa = \nabla \cdot \frac{\nabla u}{|\nabla u|}. \quad (10)$$

Due to the elastica regularizer, the resulting decision boundary of this model can have the lowest elastica energy. If set $a = 1$ and $b = 0$, this model degenerates to be the TV model. Therefore, an unified solution can be implemented for both TV model and EE model, as described in the next Section.

Here we have remarks on the curvature in high dimensional spaces. For a 1-D curve such as in image inpainting tasks, $u(x,y) = 0$ determines a level set curve according to the implicit function theorem. For a 2-D surface, the curvature given by (3) amounts to the mean curvature of this surface. From (Spivak & Spivak, 1979), the mean curvature can be defined as the average of principal curvatures. Abstractly, it can be expressed as the trace of the second fundamental form divided by the intrinsic dimension $d$. Table 1 summarizes curvature expressions in 1-D, 2-D, and high dimensional spaces. Hence the same expression(10) can be used for high dimensional situations since the constant $\frac{1}{d-1}$ can be transferred to $b$ or $\lambda$.

Table 1. Curvature expressions.

| Expression | Implicit function | Curvature |
| --- | --- | --- |
| $u(x,y) = 0$ | Planar curve $y = f(x)$ | $\kappa = \nabla \cdot \frac{\nabla u}{|\nabla u|}$ |
| $u(x,y,z) = 0$ | Surface $z = f(x,y)$ | $\kappa = \frac{1}{2}\nabla \cdot \frac{\nabla u}{|\nabla u|}$ |
| $u(x_1...x_d) = 0$ | Hypersurface $x_d = f(x_1...x_{d-1})$ | $\kappa = \frac{1}{d-1}\nabla \cdot \frac{\nabla u}{|\nabla u|}$ |

## 4. Algorithms

In contrast to discrete methods such as SVM and graph Laplacian, the proposed framework operates in a continuous fashion where powerful mathematical analysis tools can make a sense. Specifically, the calculus of variations can be exploited to minimize the energy functional, leading to the Euler-Langrange PDE.

As we have mentioned in section 3, the LR functional minimization can be transformed to solving the PDE (7). Similarly, we get the following PDE for the TV model

$$\lambda \nabla \cdot \left(\frac{\nabla u}{|\nabla u|}\right) - (u-y) = 0, \quad (11)$$

and the PDE for the EE model

$$\lambda \nabla \cdot \mathbf{V} - (u-y) = 0, \quad (12)$$

where

$$\begin{aligned}\mathbf{V} &= \phi(\kappa)\mathbf{n} - \frac{1}{|\nabla u|}\nabla(\phi'(\kappa)|\nabla u|) \\ &+ \frac{1}{|\nabla u|^3}\nabla u(\nabla u^T \nabla(\phi'(\kappa)|\nabla u|)),\end{aligned} \quad (13)$$

and $\phi(\kappa) := 1 + b\kappa^2$ by fixing $a = 1$ for simplicity. One can refer to (Chan et al., 2002) for details about the calculus of variations.

Due to the nonlinearity of the regularizer in TV and EE model, the corresponding PDE is too complicated to be efficiently solved. Even though the PDE in (7) associated with the LR model can be solved by Finite Difference Method (FDM) or Finite Element Method (FEM) in 2-D or 3-D spaces, currently we have no PDE tools to deal with high dimensional data. Therefore we take a function approximation idea by using radial basis functions (RBF), similar to the treatment in GLS (Varshney & Willsky, 2010).

### 4.1. Radial Basis Function Approximation

The function approximation idea relies on the fact that a function $u(\mathbf{x})$ can be expressed as a sum of weighted basis function $\{\varphi_i(\mathbf{x})\}$. For example, Taylor expansion represents a function by using polynomials. The most widely used is Radial Basis Function(RBF), which is simple in expressions but has powerful fitting ability. The target function $u$ can be expressed by

$$u(\mathbf{x}) = \sum_{i=1}^n w_i \varphi_i(\mathbf{x}) \quad (14)$$

with a set of Gaussian RBF

$$\varphi_i(\mathbf{x}) = \exp(-c|\mathbf{x} - \mathbf{x}_i|^2),$$

where $\{\mathbf{x}_i\}$ are the training samples in supervised learning, and $c$ is a parameter. By using RBF approximation, the problem can be reduced to finding the coefficient $\{w_i\}$.

Here are some analytical expressions that will be used later:

$$\nabla u = \sum_i w_i \nabla \varphi_i = -c \sum_i w_i(\mathbf{x} - \mathbf{x}_i)\varphi_i,$$

$$\mathbf{n} = \frac{\nabla u}{|\nabla u|} = -\frac{\mathbf{g}}{|\mathbf{g}|}, \quad \mathbf{g} := \sum_i w_i(\mathbf{x} - \mathbf{x}_i)\varphi_i,$$

$$\triangle u = \sum_i w_j \triangle \varphi_i = -c \sum_i w_i(d - c|\mathbf{x} - \mathbf{x}_i|^2)\varphi_i,$$

$$\kappa = \nabla \cdot \frac{\nabla u}{|\nabla u|} = -\frac{1}{|\nabla u|^3}\nabla u^T H(u) \nabla u + \frac{\triangle u}{|\nabla u|}$$
$$= \frac{1}{|\mathbf{g}|}\sum_i w_i \varphi_i f,$$

where $H(u)$ is the Hessian matrix of $u$, and

$$f := 1 - d + c|\mathbf{x} - \mathbf{x}_i|^2 - c\frac{\mathbf{g}^T(\mathbf{x} - \mathbf{x}_i)(\mathbf{x} - \mathbf{x}_i)^T\mathbf{g}}{\mathbf{g}^T\mathbf{g}}.$$

### 4.2. Algorithm for LR

First of all, let's consider how to deal with the LR model by solving the linear elliptic PDE (7): $-\lambda \triangle u + (u - y) = 0$. By replacing (14) into the PDE and exploiting the linearity of the Laplacian operator, the goal is to find a set of weights $\{w_i\}$:

$$\sum_i [w_i(\varphi_i - \lambda \triangle \varphi_i)] = y.$$

Let $\mathbf{w} := (w_1, w_2, ..., w_m)^T$ and $\mathbf{y} := (y_1, y_2, ...y_n)^T$, where $m$ is the number of the basis functions and $n$ is the number of the training samples. Then we have the following linear equation system:

$$\Psi \mathbf{w} = \mathbf{y}, \quad \Psi_{ij} = \varphi_j(\mathbf{x}_i) - \lambda \triangle \varphi_j(\mathbf{x}_i).$$

Numerically, the following regularized least squares solution is considered in practise to avoid ill-posed problems:

$$\min_\mathbf{w} |\Psi \mathbf{w} - \mathbf{y}|^2 + \eta |\mathbf{w}|^2.$$

The solution is simply given by $\mathbf{w} = (\Psi^T\Psi + \eta I)^{-1}\Psi^T\mathbf{y}$ with a very fast speed.

### 4.3. Algorithm for TV and EE models

As the TV model is one special case of the EE model, we describe solutions for the more complicated EE model in this section. Here two algorithms are developed to tackle the nonlinearity: (1) gradient Descent time marching, and (2) Lag-Linear Equation iteration.

#### 4.3.1. Gradient Descent Time Marching

Using the calculus of variations, we can get the descent gradient for the desired function. With a matrix notation $u(\mathbf{x}) = \Phi \mathbf{w}$ where $\Phi_{ij} = \varphi_j(\mathbf{x}_i)$, the gradient of $u$ is given as follows when $\Phi$ is fixed:

$$\Phi \frac{\partial \mathbf{w}}{\partial t} = \begin{pmatrix} \frac{\partial u}{\partial t}|_{\mathbf{x}=\mathbf{x}_1} \\ \vdots \\ \frac{\partial u}{\partial t}|_{\mathbf{x}=\mathbf{x}_n} \end{pmatrix}.$$

Then each iteration can be written as

$$\mathbf{w}^{(k+1)} = \mathbf{w}^{(k)} - \tau \frac{\partial \mathbf{w}}{\partial t} = \mathbf{w}^{(k)} - \tau \Phi^{-1} \begin{pmatrix} \frac{\partial u^{(k)}}{\partial t}|_{\mathbf{x}=\mathbf{x}_1} \\ \vdots \\ \frac{\partial u^{(k)}}{\partial t}|_{\mathbf{x}=\mathbf{x}_n} \end{pmatrix},$$

where $\tau$ is a small time step, and $u^{(k)}$ renews from $\mathbf{w}^{(k)}$. The coefficients $\mathbf{w}$ is initialized as $\mathbf{w}^{(0)} = (\Phi^T \Phi + \eta I)^{-1}\Phi^T \mathbf{y}$. Then the gradient $\frac{\partial u}{\partial t}$ must be figured out first.

From the PDE (11), the gradient of $u$ for the TV model is simply given by:

$$\frac{\partial u}{\partial t} = \lambda \nabla \cdot \left(\frac{\nabla u}{|\nabla u|}\right) - (u - y).$$

From (12) and (13), the gradient of $u$ for the EE model can be written as:

$$\frac{\partial u}{\partial t} = \lambda \nabla \cdot \mathbf{V} - (u - y).$$

Through several steps of calculations by leaving out three and higher order terms, $\nabla \cdot \mathbf{V}$ can be expanded into the following expression:

$$\nabla \cdot \mathbf{V} = \kappa - \frac{4b\kappa\triangle u}{|\nabla u|^4}\nabla u^T H(u) \nabla u + b\kappa^3 + 2b\Big(\frac{2\triangle u}{|\nabla u|^3} + \frac{\kappa}{|\nabla u|^4}\Big)(\nabla u^T H(u))(\nabla u^T H(u))^T + 2b\Big\{\frac{\triangle u}{|\nabla u|^3} - \frac{3}{|\nabla u|^5}\nabla u^T H(u)\nabla u\Big\}\Big(-\frac{2\triangle u}{|\nabla u|^2} + \frac{\kappa}{|\nabla u|}\Big)\nabla u^T H(u)\nabla u,$$

where $H(u)$ is the Hessian matrix of $u$, and

$$\kappa = \nabla \cdot \frac{\nabla u}{|\nabla u|} = \frac{\triangle u}{|\nabla u|} - \frac{1}{|\nabla u|^3}\nabla u^T H(u) \nabla u.$$

We can see that if setting $b = 0$ the expression is degraded to $\nabla \cdot \mathbf{V} = \kappa = \nabla \cdot \frac{\nabla u}{|\nabla u|}$, which is exactly the same expression for the TV model. The time complexity in each iteration is $O(n^2 d)$, where $n$ is the number of data points and $d$ is the dimension. We set the maximal number of iterations as 40. There are 3 parameters in the algorithm: the RBF parameter $c$, the regularization parameter $\lambda$, and the elastica weight parameter $b$. Note that we set $a = 1$ since $a$ can be absorbed into $\lambda$.

4.3.2. LAGGED LINEAR EQUATION ITERATION

Following the spirit of the lagged diffusivity fixed-point iteration method in (Chan & Shen, 2005), we develop the following lagged linear equation iteration method. Empirically, the original lagged diffusivity fixed-point iteration often yields poor performance due to its brute-force linearization on the nonlinear PDE.

For the simpler TV model, by expanding the nonlinear term $\nabla \cdot (\nabla u/|\nabla u|)$ we have

$$-\frac{\lambda}{|\nabla u|}(\triangle u - \frac{\nabla u^T H(u) \nabla u}{\nabla u^T \nabla u}) + (u - y) = 0,$$

where $H(u)$ is the Hessian of function $u$. Then plugging the RBF approximation into above PDE, we get the following system

$$\sum_i w_i(\frac{|\mathbf{g}|}{\lambda} - f)\varphi_i = |\mathbf{g}|y, \quad (15)$$

where

$$\mathbf{g} := \sum_i w_i(\mathbf{x} - \mathbf{x}_i)\varphi_i,$$

$$f := 1 - d + c|\mathbf{x} - \mathbf{x}_i|^2 - c\frac{\mathbf{g}^T(\mathbf{x} - \mathbf{x}_i)(\mathbf{x} - \mathbf{x}_i)^T \mathbf{g}}{\mathbf{g}^T \mathbf{g}},$$

and $d$ is the data dimension. Using the lagged idea, we obtain the lagged linear equation iteration algorithm: 1) By fixing $\mathbf{g}$, solve the system of linear equations with respect to $\mathbf{w}$ to get a new $\mathbf{w}$; 2) Compute $\mathbf{g}$ with updated $\mathbf{w}$; 3) Iterate until convergence or maximal iteration number.

For the more complicated EE model, we have

$$\sum_i w_i(\frac{|\mathbf{g}|}{\lambda K} - f)\varphi_i = |\mathbf{g}|y, \quad (16)$$

where

$$K := a + b\kappa^2 = a + b\big(\frac{1}{|\mathbf{g}|}\sum_i w_i \varphi_i f\big)^2.$$

Similarly, a two-step lagged iteration procedure can be developed for the EE model: 1) By fixing $\mathbf{g}$ and $K$, solve the linear system with respect to $\mathbf{w}$; 2) Compute $\mathbf{g}$ and $K$ with updated $\mathbf{w}$; 3) Iterate until convergence or maximal iteration number. There are three parameters: $c$, $\lambda$, and regularization parameter $\eta$ (empirically chosen in experiments) in the least squares problems.

## 5. Experimental Results

The proposed two approaches (TV and EE) are compared with LR, SVM (with RBF kernels), and Back-Propagation Neural Networks (BPNN) for binary classification, multi-class classification, and regression on benchmark data sets.

### 5.1. Binary Classification

The test data sets for binary classification are from the libsvm website. Originally, these data sets are scaled to [0,1] and serve as benchmark to test the libsvm implementation. Here we downloaded seven data sets to evaluate performance of our methods (TV and EE) with two kinds of implementations of Gradient Descent method (GD) and Lagged Linear Equation method (lagLE).

The optimal parameters for each algorithm are selected by grid search using 5-fold cross-validation. To make the grid search more practical, only the two common parameters ($c$ and $\lambda$) are searched for SVM, LR, TV, and EE except BPNN. Empirically, the parameter $\eta$ is set as 1 for LR, and the parameter $b$ is fixed as 0.01 for EE. Then excluding BPNN, the two common parameters are searched from logarithm from $-10:10$ with step 2. For each data set, we randomly run the 5-fold cross validation ten times to reduce the influence of data partition. Table 2 shows the average classification accuracies for the five methods.

From the table we can see that BPNN performs worst, while the LagLE solution of EE outperforms others on 5 data sets. The GD implementation of TV and EE is competitive with SVM. And similar accuracies are achieved by TV and EE partially because of their close connections.

### 5.2. Multi-class Classification

For multi-class tests, we collected data sets from libsvm website and UCI Machine Learning Repository, including frequently used small data sets and the USPS handwritten digital set. For USPS data, PCA is used to reduce the dimension to 30 and we randomly select 1000 samples for experiments.

Except for BPNN that has a built-in ability for multi-class tasks, almost all function learning approaches are originally designed for binary classification. In order to handle multi-class situations, usually one versus all or one versus one strategies can be adopted. If using one vs all, one needs to learn $M$ functions to fulfill the multi-class task, where $M$ is the number of classes. Recently in (Varshney & Willsky, 2010), an efficient binary encoding strategy was proposed to represent the decision boundary by only $log_2 M$ functions. In our experiments the one vs all strategy is used. Same as the binary problems, we use the 5-fold cross-validation to choose the optimal parameters for each method. Except BPNN, all other methods have 2 common parameters which are searched from logarithm from $-10:10$ with step 1. The results of average

Table 2. Average accuracies (%) for binary classification using 5-fold cross-validation.

| Data | Dim | Num | SVM | BPNN | LR | TV | | EE | |
|---|---|---|---|---|---|---|---|---|---|
| | | | | | | GD | lagLE | GD | lagLE |
| liver-disorders | 6 | 345 | 73.96 | 71.52 | 73.20 | 74.81 | 73.62 | 74.32 | 73.91 |
| diabetes | 8 | 768 | 78.07 | 76.85 | 77.96 | 77.50 | 77.81 | 77.23 | 78.10 |
| breast-cancer | 10 | 683 | 97.22 | 96.23 | 97.60 | 97.13 | 97.72 | 97.13 | 97.83 |
| heart | 13 | 270 | 84.85 | 81.76 | 84.26 | 80.05 | 84.58 | 80.00 | 84.96 |
| australian | 14 | 690 | 86.41 | 86.34 | 87.04 | 86.99 | 87.01 | 86.54 | 87.10 |
| german-number | 24 | 1000 | 76.94 | 74.16 | 77.10 | 76.19 | 77.10 | 76.50 | 77.22 |
| sonar | 60 | 208 | 88.80 | 82.99 | 90.88 | 90.30 | 89.27 | 90.07 | 90.50 |

Table 3. Average accuracies (%) for multi-class classification using 5-fold cross-validation.

| Data | Classes | Dim | Num | SVM | BPNN | LR | TV | | EE | |
|---|---|---|---|---|---|---|---|---|---|---|
| | | | | | | | GD | lagLE | GD | lagLE |
| iris | 3 | 4 | 150 | 96.00 | 96.00 | 95.33 | 96.00 | 96.00 | 96.00 | 96.00 |
| balance | 3 | 4 | 625 | 99.68 | 92.48 | 89.44 | 90.88 | 89.92 | 90.40 | 90.01 |
| hayes | 3 | 5 | 132 | 80.30 | 74.26 | 71.57 | 77.87 | 73.08 | 77.87 | 76.15 |
| tae | 3 | 5 | 151 | 62.25 | 56.63 | 59.47 | 64.18 | 66.00 | 61.41 | 66.00 |
| wine | 3 | 13 | 178 | 99.44 | 97.78 | 99.44 | 99.44 | 99.43 | 99.44 | 98.86 |
| vehicle | 4 | 18 | 846 | 85.70 | 79.18 | 82.75 | 85.00 | 82.25 | 85.00 | 82.84 |
| glass | 6 | 9 | 214 | 72.43 | 63.99 | 73.81 | 69.59 | 76.19 | 67.72 | 75.71 |
| segment | 7 | 19 | 500 | 92.40 | 90.60 | 91.80 | 90.80 | 93.55 | 91.20 | 95.89 |
| flag | 8 | 29 | 194 | 52.06 | 46.90 | 53.13 | 49.50 | 52.10 | 50.55 | 52.10 |
| yeast | 10 | 8 | 1484 | 61.19 | 54.49 | 58.22 | 57.95 | 57.91 | 57.95 | 57.97 |
| usps | 10 | 30 | 1000 | 93.90 | 82.60 | 94.90 | 94.40 | 94.80 | 94.40 | 95.00 |

accuracies are shown in table 3.

The accuracy results demonstrate that BPNN performs the worst, while TV and EE are comparable to SVM on the 11 test data sets. Compared with TV/EE, SVM achieves higher accuracies on 4 data sets, performs worse on 4 data sets, and offers the same best accuracies on 2 data sets. One reason might be that very complex and even wiggly decision boundaries are preferred for some multi-class data sets. Due to the strong regularization on the geometric shapes, TV/EE can not adapt to yield complex decision hypersurfaces for these data sets.

### 5.3. Regression

We use seven regression data sets from UCI Machine Learning Repository to validate the proposed methods compared with SVM, BPNN, and LR. All data sets are scaled to [0,1]. Note that here the Gradient Descent (GD) method is used for TV and EE. We take the same experimental settings by running ten times of 5-fold cross-validation for each data set. Table 4 shows the regression results using mean square errors (MSE). Clearly, we can see that both TV and EE achieve lower MSE than SVM and LR on 6 data sets. Also TV and EE outperform BPNN on 5 data sets. The results demonstrated superb regression ability of our proposed methods.

### 6. Conclusion

Regularization framework and function learning approaches have become very popular in the recent machine learning literature. Due to the great success of total variation and Euler's elastica models in image processing area, we extend these two models for supervised classification and regression on high dimensional data sets. The TV regularizer permits steeper edges near the decision boundaries, while the elastica smoothing term penalizes non-smooth level set hypersurfaces of the target function. Compared with SVM and BPNN, our proposed methods have demonstrated the competitive performance on commonly used benchmark data sets. Specifically, TV and EE models achieve better performance on most data sets for binary classification and regression. Currently one main disadvantage is the slow convergence speed of iteration procedures. The future work is to explore other fitting loss term such as hinge loss, to study other possibilities of basis functions, to investigate the existence and uniqueness of the PDE solutions, and to reduce the running time.

Table 4. Regression results measured by MSE ($10^{-3}$) using 5-fold cross-validation.

| Data | Dim | Num | SVM | BPNN | LR | TV | EE |
|---|---|---|---|---|---|---|---|
| SERVO | 4 | 167 | 10.856 | 5.623 | 7.290 | 8.339 | 7.860 |
| MACHINECPU | 6 | 209 | 3.341 | 5.175 | 1.782 | 1.907 | 1.754 |
| AUTOMPG | 7 | 392 | 6.958 | 5.633 | 6.072 | 5.620 | 5.686 |
| CONCRETE | 8 | 1030 | 6.124 | 4.884 | 6.019 | 5.432 | 5.236 |
| HOUSING | 13 | 506 | 6.200 | 7.540 | 5.130 | 4.897 | 4.951 |
| PYRIM | 27 | 74 | 9.911 | 23.058 | 6.590 | 5.766 | 6.005 |
| TRIAZINES | 60 | 186 | 19.712 | 41.902 | 20.734 | 20.515 | 20.947 |

## Acknowledgments

The authors would like to thank the anonymous reviewers for their helpful suggestions. This work was supported by the National Basic Research Program of China (973 Program number 2011CB302202) and the National Science Foundation of China (NSFC grant 61075119).